%% file: main.tex
\documentclass[sigconf]{acmart}

\usepackage{colortbl}
\usepackage{booktabs} 
\usepackage{balance}

\usepackage[ruled,linesnumbered]{algorithm2e}

\newtheorem{mydef}{Definition}

\usepackage{graphicx}
\usepackage{subcaption}
\usepackage{textcomp}
\usepackage{xcolor}
\usepackage{multirow}
\newcommand{\argmin}{\mathop{\mathrm{argmin}}}   

\begin{document}
\title{TestRank: Bringing Order into Unlabeled Test Instances for Deep Learning Tasks } 

\author{Yu Li$^\dagger$, Min Li$^\dagger$, Qiuxia Lai$^\dagger$, Yannan Liu$^*$, and Qiang Xu$^\dagger$}

\affiliation{
  \institution{
  $^\dagger$ The Chinese University of Hong Kong, Shatin, N.T., Hong Kong \\
  $^*$ Independent Researcher \\
  Email: $^\dagger$ \{yuli, mli, qxlai, qxu\}@cse.cuhk.edu.hk }
  \country{}
}
\renewcommand{\authors}{Yu Li, Min Li, Qiuxia Lai, Yannan Liu and Qiang Xu}

\begin{abstract}
Deep Learning (DL) models have achieved unprecedented success in various tasks and are pervasively deployed in real-world applications. It is critical to guarantee their correctness by testing their behaviors. However, DL systems are notoriously difficult to test and debug due to the lack of explainability and the huge test input space to cover. Generally speaking, it is relatively easy to collect a massive amount of test data, but the labeling cost can be quite high. Consequently, it is essential to conduct test selection and label only those selected `high quality' bug-revealing test inputs for test cost reduction.

In this paper, we propose a novel test prioritization technique that brings order into the unlabeled test instances according to their bug-revealing capabilities, namely~\textit{TestRank}. Different from existing solutions, TestRank leverages both \emph{intrinsic} attributes and \emph{contextual} attributes of test instances when prioritizing them. 
To be specific, we first build a similarity graph on test instances and training samples, and we conduct graph-based semi-supervised learning to extract contextual features. 
Then, for a particular test instance, the contextual features extracted from the graph neural network (GNN) and the intrinsic features obtained with the DL model itself are combined to predict its bug-revealing probability. Finally, TestRank prioritizes unlabeled test instances in descending order of the above probability value.
We evaluate the performance of TestRank on a variety of image classification datasets. Experimental results show that the debugging efficiency of our method significantly outperforms existing test prioritization techniques. 

\end{abstract}



\maketitle

\input{intro} 
\input{background} 
\input{method} 
\input{result} 
\input{conclusion}
\balance
\bibliographystyle{ACM-Reference-Format}
\bibliography{ref}

\end{document}

%% file: intro.tex
\section{Introduction}

Deep Learning (DL) has become the de facto technique in artificial intelligence systems as it achieves state-of-the-art performance in various tasks. Despite their effectiveness, on the one hand, DL models are lack of explainability and hence cannot be proved to be correct; on the other hand, they are prone to errors due to many factors, such as the biased training/validation data, the limitations of the model architecture, and the constraints on training cost. 
Needless to say, it is essential to conduct a high-quality test and debug before DL models are deployed in the field, otherwise the behaviors of DL models can be unpredictable and result in accidents after deployment. 
Traditional machine learning models are tested on personally-collected test instances with manual labels. Nowadays, due to the easy access to online resources, the test data is ubiquitous but most of them are unlabeled ones. The cost of building test oracles by manually labeling a massive set of test instances is prohibitive, especially for tasks requiring experts for accurate labeling, such as medical images and malware executables. 

 \begin{figure}[t]
     \centering
     \includegraphics[width=\linewidth]{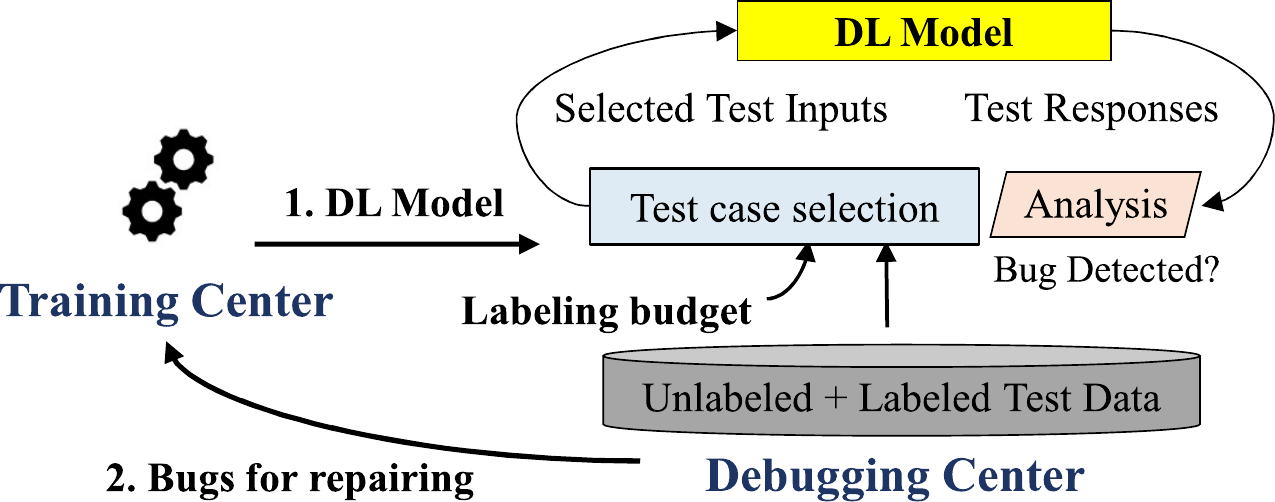}
     \caption{The debugging process for deep learning models.}
     \label{figs:debug}
 \end{figure}



To tackle the above problem, the objective of test prioritization~\cite{Bogdan2007ModelTest, Rajput2020TechniquesOT, Elbaum2001IncorporatingVT} is to identify a set of `high-quality' test instances, which facilitates to reveal more bugs of the DL model with less labeling effort and hence significantly reduces the test cost. The general test flow of DL models with test prioritization is depicted in Figure \ref{figs:debug}. 
For a DL model trained by the training center, the debugging center will first conduct test prioritization to select those test inputs from the vast unlabeled data pool under certain labeling budgets (e.g. the number of instances to be labeled). The test responses of the selected test instances are compared with their oracle labels, and the corresponding bug reports can be sent to the training center for model repair.

Several test prioritization techniques have been proposed for DL models. 
These methods try to derive the bug-revealing capability for a test instance based on the estimated distance to the decision boundary with its \textit{intrinsic attributes} (\textit{e.g.}, the softmax-based probabilities given by the target DL model to this specific input). 
DeepGini \cite{Feng2020DeepGini} proposes to feed the unlabeled data to the target DL model and calculate confidence-related scores based on the model's output probabilities to rank the unlabeled test cases. Test cases with nearly equal probabilities on all output classes are regarded as less confident ones, and hence likely to reveal model bugs. Similarly, Byun \textit{et. al.} propose to use the uncertainty score from the target DL model for test input prioritization~\cite{ByunSVRC19}. Multiple-boundary clustering and prioritization (MCP) \cite{Shen2020MCP} considers both the output probabilities and the balance among each classification boundary when selecting test cases.
However, near-boundary instances are not necessarily bugs, especially for well-trained classifiers with high accuracy. 
Also, as bugs can be far from the decision boundary, methods that only utilize such distance estimation can never reveal all buggy instances.

To estimate a test instance's capability in triggering bugs, besides the intrinsic attributes mentioned above, there is another type of information that provides extra insight into the target DL model's behavior: the known classification correctness of labeled samples (i.e., training samples and test oracles) from the target DL model. Such historical data is easily accessible and it provides contextual information that reflects the distribution of the high-dimensional data and their corresponding inference correctness with the target model.

In this work, we present a novel test prioritization technique, namely \textit{TestRank}, which exploits both the intrinsic and the contextual attributes of test instances to evaluate their bug-revealing capabilities. 
As similar inputs are usually associated with the same labels, we propose to construct a similarity graph on both unlabeled and the historical labeled instances to extract the contextual attributes of unlabeled test instances with graph neural networks (GNNs). After that, we aggregate intrinsic and contextual attributes with a neural-network-based binary classifier for test prioritization. The main contributions of this work include:
 \begin{itemize}
 	\item To the best of our knowledge, \textit{TestRank} is the first work that takes the contextual attributes of the target DL model into consideration for test prioritization.
 	\item We propose to construct a similarity graph on both labeled and unlabeled samples, and construct a graph neural network to learn to extract useful contextual attributes for these unlabeled instances. We also present approximation techniques to reduce its computational complexity with minor impact on the performance of \textit{TestRank}. 
 	\item We propose a simple yet effective neural network that combines the intrinsic attributes and the contextual attributes of unlabeled test instances for their bug-revealing capability estimation. 
 \end{itemize}

We evaluate the proposed \textit{TestRank} technique on three popular image classification datasets: CIFAR-10, SVHN, and STL10. The results show that our method outperforms the state-of-the-art methods by a considerable margin. 

The rest of the paper is organized as follows. In Section \ref{sec:background}, we present the necessary background about the test prioritization problem and the graph neural network algorithms. We illustrate our problem formulation and motivation in Section \ref{sec:prob}. Then, Section \ref{sec:metho} details the proposed \textit{TestRank} solution. Experimental results are presented in Section \ref{sec:experiment}. Finally, Section \ref{sec:conclusion} concludes this paper.







%% file: background.tex
\section{Background}
\label{sec:background}
In this section, we first introduce existing test prioritization techniques for the DL model. Then, we present the necessary background on graph neural network algorithms, which is used for extracting the contextual attributes.

\subsection{Test Prioritization for DL Model}
\label{sec:test_prioritization}

For DL models, two categories of criteria have been proposed to qualify the test instance during test prioritization. One is \textit{internal-state-based}~\cite{pei2017deepxplore, Ma2018DeepGaugeMT, Kim2019GuidingDL}, which defines the quality on DL model's internal status, such as the activation values in hidden layers. The other one is \textit{final-output-based}~\cite{Feng2020DeepGini,Shen2020MCP}, which examines the final output of DL model.

\vspace{3pt}
\noindent
\textbf{Internal State Based.} 
Similar to the code coverage in traditional software testing~\cite{zhang2018predictive}, multiple versions of coverage metrics are proposed to measure the degree to which the DNN's internal neurons are exercised by the test set. The selected set of test instances with a higher coverage value is believed with higher quality for test prioritization. DeepXplore~\cite{pei2017deepxplore} proposes the first Neuron Activation Coverage (NAC) metric, which measures the fraction of neurons whose output values are larger than a predefined threshold. 
Inspired by DeepXplore, DeepGauge \cite{Ma2018DeepGaugeMT} refines the criteria by segmenting the activation value of each neuron into multiple sections, such as k-Multisection Neuron Coverage (KMNC), Neuron Boundary Coverage (NBC),  and so on. The coverage is thereby defined as the percentage of exercised segments over all possible segments. 


Besides these neuron activation coverage metrics, surprise coverage further enhances the NAC by diversifying test instances from training instances. Kim \textit{et.al.} proposes two kinds of surprise criteria~\cite{Kim2019GuidingDL}: the likelihood of an instance has been seen during training via Likelihood-based Surprise Adequacy Coverage (LSA) and the distance of an instance to the training instances via Distance-based Surprise Adequacy Coverage (DSA). To be specific, LSA uses kernel density estimation to obtain a density function from the training instances. The resulting density function allows the estimation of the likelihood of a new input instance. For a test instance $x$, DSA first calculates the $dist_a$, which is the distance between the activation trace of $x$, denoted by $\alpha_\mathbf{N}(x)$, and the closest trace $\alpha_\mathbf{N}(x_a)$ produced by sample $x_a$ in the training set $\mathbf{D}$ with the same predicted class $C_x$ as $x$.
Then, DSA calculates $dist_b$, which is the distance between the $\alpha_\mathbf{N}(x_a)$ and the closest trace $\alpha_\mathbf{N}(x_b)$ in the training set with different predicted class as sample $a$. The final distance is then defined as $\frac{dist_a}{dist_b}$:
\begin{equation}
    dist_a = \|\alpha_\mathbf{N}(x) - \alpha_\mathbf{N}(x_a)\|
\end{equation}
\begin{equation}
    x_a = \argmin_{\mathbf{D}(x_i) = c_x}\|\alpha_\mathbf{N}(x) - \alpha_\mathbf{N}(x_i)\|
\end{equation}
\begin{equation}
            dist_b = \|\alpha_\mathbf{N}(x_a) - \alpha_\mathbf{N}(x_b)\|   
\end{equation}
\begin{equation}
    x_b = \argmin_{\mathbf{D}(x_i) \in C \setminus \{c_x\}}\|\alpha_\mathbf{N}(x_a) - \alpha_\mathbf{N}(x_i)\|,
\end{equation}
where the activation traces can be the output from the intermediate layers of a DL model. 

However, all the above coverage-based criteria are inefficient in recalling the bug-trigger test instances for test prioritization ~\cite{Feng2020DeepGini}. Even worse, neuron activation coverage metrics can be negatively correlated with bug detection capability~\cite{Yan2020Correlation, HarelCanada2020IsNC}.

\begin{figure}
	\includegraphics[width=\linewidth]{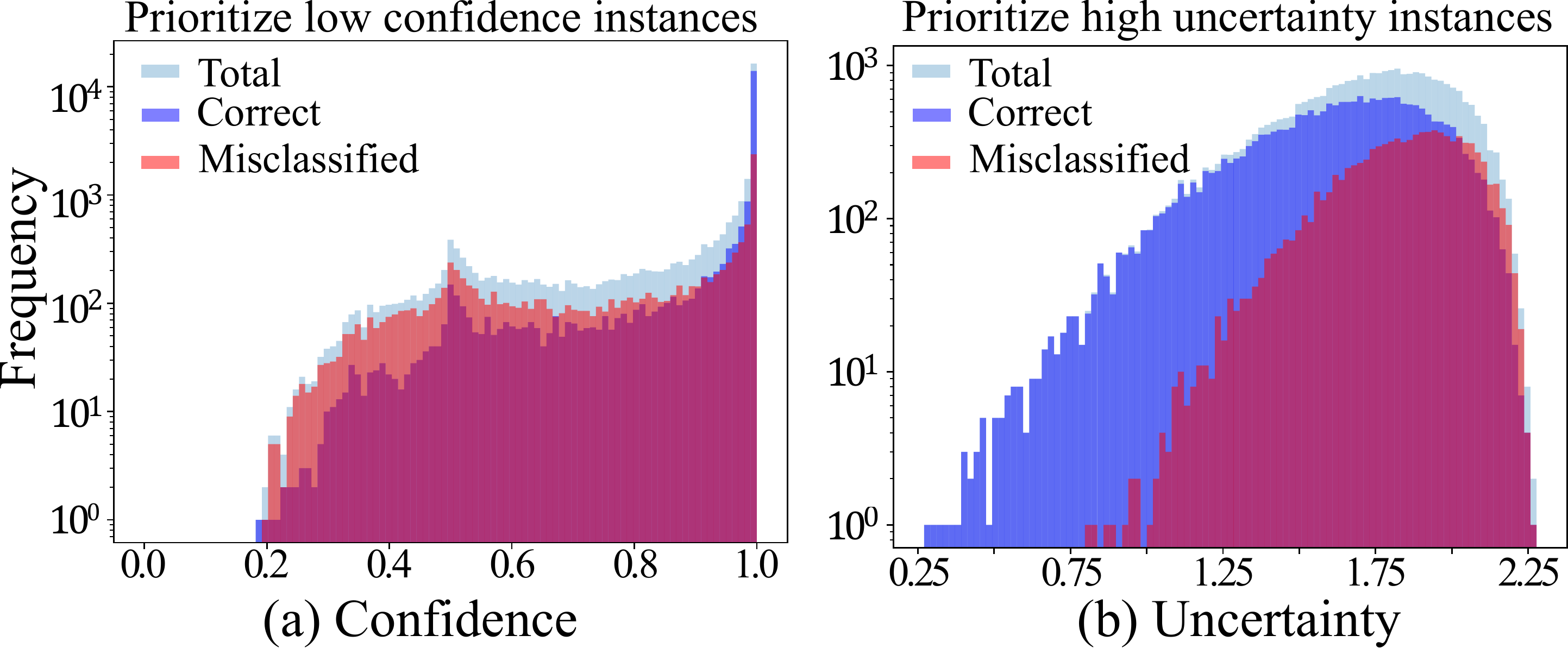}
	\caption{The histogram of a) Confidence scores b) Dropout uncertainty of correctly classified and misclassified test instances. This result is obtained from a DNN classifier trained on the CIFAR-10 dataset. Note that the purple area is the overlap between the red and blue areas. As we can observe, many correctly classified test instances have low confidence scores in (a) or high uncertainty scores in (b), which means that these metrics cannot accurately measure the bug-revealing ability of test instances.}
	\label{fig:confidence}
\end{figure}

\vspace{3pt}
\noindent
\textbf{Final Output Based.} \label{sec:final_output_based}
Instead of examining the DL model's internal status, DeepGini~\cite{Feng2020DeepGini} proposes to qualify a single test instance via the DL model's final output. If a DL model predicts a test instance with nearly equal probability values on all candidate classes, the prediction result is less confident and likely to be an incorrect prediction. Formally, DeepGini measures the likelihood of incorrect prediction of a test instance $t$ by the confidence of final output, dedicated by:
\begin{equation}
    \label{eq:confidence}
	confidence(t) = 1 - \Sigma_{i=1}^N p_{t,i}^2, 
\end{equation}
where, $p_{t,i}$ is the predicted probability that the test case $t$ belongs to the class $i$ by DNN. Given the sum of $p_{t,i}$ is 1, $confidence(t)$ is minimized when all $p_{t,i}$ values are equal, and test instance with lower confidence score is of higher quality during testing prioritization.
Instead of evaluating the overall likelihood for all classes, Multiple-Boundary Clustering and Prioritization (MCP) proposes to evaluate the likelihood of misclassification for each pair of classes individually~\cite{Shen2020MCP}. In this way, test instances can be evenly selected for each class pair and the misclassification cases are investigated at the finer granularity. Besides these metrics, Byun \textit{et. al.} also propose to measure the likelihood of incorrect prediction by the uncertainty of the final output~\cite{ByunSVRC19}, which reflects the degree to which a model is uncertain about its prediction. In practice, evaluating uncertainty requires the task DL model designed in a Bayesian Neural Network form~\cite{richard1991neural,neal2012bayesian} or containing dropout layer for approximation~\cite{GalG16}.

We regard the output given by the DL model to an input instance as the intrinsic attributes of this input. However, the criteria based on the intrinsic attributes are also inefficient in recognizing bug-triggering test instances. We take the confidence \cite{Feng2020DeepGini} and uncertainty \cite{ByunSVRC19} metrics as an example. Figure~\ref{fig:confidence}(a) shows the distribution of CIFAR-10 test instances over the confidence values. We observe that the confidence values of most misclassified test instances tend to be high, where the test instances for correct classification and misclassification are not distinguishable. Figure~\ref{fig:confidence}(b) shows the distribution over uncertainty, and the observation is similar to the confidence case, that correct classification and misclassification are not distinguishable in most cases.


\subsection{Graph Neural Network}
Besides the above intrinsic attributes used by current test prioritization techniques, in this paper, we propose to extract contextual attributes from the contextual information (\textit{e.g.}, the historical inputs and classification correctness from the DL model).

Graph-based representation learning is a way to extract contextual attributes from the contextual information \cite{perozzi2014deepwalk, KipfW17}. 
Given a graph $G = (\mathbf{V}, \mathbf{E}, \mathbf{W})$, where $\mathbf{V}$ is the node set, $\mathbf{E}$ is the edge set, and $\mathbf{W}$ is the edge weight matrix.
The underlying goal of graph-based representation learning is to learn a representation for each node in the graph based on its context (e.g. neighbor's node features).

Graph neural networks (GNN) are powerful techniques that have been proposed for graph-based semi-supervised representation learning~\cite{KipfW17,Xu2019HowPA,Velickovic2018GraphAN, Hamilton2017InductiveRL, Rong2020DropEdgeTD}. 
Kipf \textit{et. al.}~\cite{KipfW17} are the first one to propose the message-passing graph convolutional network (GCN).
To be specific, they define a layer-wise information aggregation and propagation rule, which is given by 
\begin{equation}
    \mathbf{H}^{l+1} = \sigma(\mathbf{\tilde{D}}^{-\frac{1}{2}}\mathbf{\tilde{A}\tilde{D}}^{-\frac{1}{2}}\mathbf{H}^{l}\mathbf{\Theta}^{l}),
\end{equation}
where $\mathbf{\tilde{A}} = \mathbf{A} + \mathbf{I}_{N}$ is the adjacency matrix of $G$ with added self-connections. Note that $\mathbf{A}$ can include other values than 1 to represent the edge weights.  $\mathbf{I}_{N}$ is the identity matrix, $\mathbf{\tilde{D}} = \sum_j{\mathbf{\tilde{A}}_{i,j}}$, $\mathbf{\Theta^{l}}$ is the trainable weight matrix for the $l^{th}$ layer, $\sigma$ is an activation function (e.g. ReLU, Tanh, etc) and $H^l$ is the representation matrix of the $l^{th}$ layer. 
Then, a neural network model based on graph convolutions can be built by stacking multiple convolutional layers. For example, a two-layer GNN model would be:
\begin{equation}
 \mathbf{H}^{2} = \sigma(\mathbf{A}^{'}\sigma(\mathbf{A}^{'}\mathbf{H}^{0}\mathbf{\Theta}^{0})\mathbf{\Theta}^{1}),
\end{equation}
where $\mathbf{H}^0$ is the input node feature matrix, $\mathbf{H}^{2}$ is the output node representation matrix, and $\mathbf{A}^{'} = \mathbf{\tilde{D}}^{-\frac{1}{2}}\mathbf{\tilde{A}}\mathbf{\tilde{D}}^{-\frac{1}{2}}$. 
The graph neural networks are then trained on the labeled data on the graph with certain loss functions such as cross-entropy loss. 
Later, the trained model can be applied to the unlabeled data for representation extracting.

The follow up works extend the GNN by considering diverse aggregation functions \cite{Hamilton2017InductiveRL}, introducing attentions mechanism to GNN \cite{Velickovic2018GraphAN}, and so on.

%% file: method.tex
\section{The Problem and Motivation}
\label{sec:prob}
In this section, we formally define the problem in Section \ref{subsec:problem}. Then, we provide a motivational example in Section \ref{subsec:motivation}.

\begin{figure}
	\includegraphics[width=\linewidth]{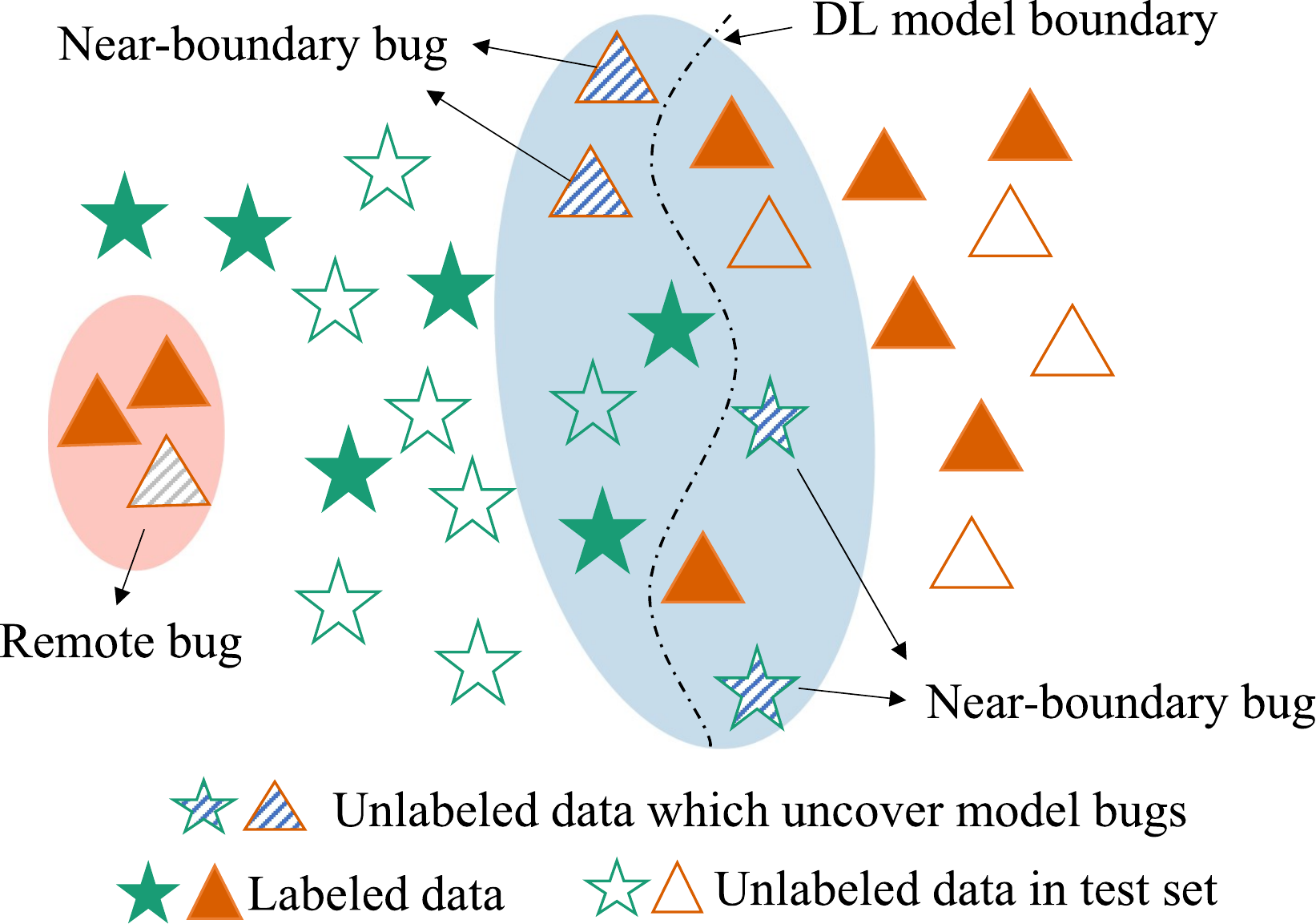}
	\caption{A motivational example. Existing approaches aim to select near-boundary instances since they assume these instances are likely to be misclassified and hence reveal potential bugs. However, near-boundary instances are not necessarily bugs. What's worse, they ignore those bugs that are far from the decision boundary, i.e., remote bugs. To enhance the detection of both near-boundary and remote bugs, TestRank makes use of the relation between the unlabeled instances and labeled samples.}
	\label{fig:motivation}
\end{figure}

\subsection{Problem Formulation}
\label{subsec:problem}


Let us use $f: \mathcal{X} \to \mathcal{Y}$ to represent the given target DL model, where $\mathcal{X}$ and $\mathcal{Y}$ are the input and output space, respectively. the debugging center is required to select a certain number of test instances from the unlabeled test instance pool that can trigger model bugs as many as possible.
Later, these triggered bugs can be fed back to the training center for repairing or conduct bug analysis. 
We define the wrong behaviors as bugs for DL models as follows:

\begin{mydef}
\textbf{Bug of DL Models.} A bug of the DL model is a defect of the model and can be uncovered by the test instance $\mathbf{x}$ if the predicted label $f(\mathbf{x})$ is inconsistent with its ground truth label $y_\mathbf{x}$, namely $f(\mathbf{x}) \neq y_\mathbf{x}$.
\end{mydef}

Specifically, to conduct an effective testing, the debugging center selects $b$ test cases (the set of selected test cases is denoted as $\mathbf{X}_S$) from the unlabeled test instance pool $\mathbf{X}_U$ to identify as many as bugs in DL model $f$. Therefore, the objective is to maximize the cardinality of set $\lbrace \mathbf{x}|f(\mathbf{x}) \neq y_x \rbrace$ under the budget constraint:
\begin{equation}
    \max |\lbrace \mathbf{x}|f(\mathbf{x}) \neq y_\mathbf{x} \rbrace|, \text{where } \mathbf{x} \in \mathbf{X}_S\text{ and }|\mathbf{X}_S| = b.
\end{equation}

\subsection{Motivation}
\label{subsec:motivation}
Intuitively, the bug-revealing capability of an unlabeled test input is closely related to the input instance's attributes for a specific model to be tested. In this work, we distinguish two kinds of attributes for an unlabeled instance: the \textit{intrinsic attributes} and the \textit{contextual attributes}. 
The \textit{intrinsic attributes} of input are defined as the output responses assigned by the target DL model to this input. This could be for example the output predictive distribution for the input from the target DL model,
reflecting the probability of the input belonging to each class.
This kind of attribute is used by existing test input prioritization approaches \cite{Feng2020DeepGini, Shen2020MCP, ByunSVRC19}. 
Besides the intrinsic attributes, the contextual attributes provide a deeper insight on the target model behavior: 
the historically-exercised labeled inputs and their classification correctness.
For a particular test instance, such contextual attributes are complementary to its intrinsic attributes, which should be combined to tell its bug-revealing capability.

An illustrative example is shown in Figure \ref{fig:motivation}, wherein we visualize the behavior of a two-class classifier on the unlabeled test data and historically labeled data distribution. The blue region includes the instances that are near the decision boundary. Intuitively, the classifier is uncertain about the data when data is near the decision boundary and thus is likely to misclassify it.
Existing works~\cite{Shen2020MCP, Feng2020DeepGini, ByunSVRC19} propose various solutions to look for near-boundary instances to be selected, \textit{e.g.}, metrics like confidence/uncertainty/surprise scores to qualify the distance from the data to the decision boundary. 
However, the near-boundary instances are not necessarily bugs and most of them can be correctly classified for a well-trained classifier. 
To make things worse, such testing approaches fail to consider the bugs lying far from the decision boundary (\textit{i.e.}, remote bugs, shown in the red region in Figure~\ref{fig:motivation}), because DL models usually output high confidence (or low uncertainty) for these inputs.
These bugs may result from the limited model capacity, insufficient training data, or other factors. 

Our key insight is that we can use the contextual information to help locate the near-boundary bugs as well as remote bugs.
The usefulness of the contextual information is due to the local continuity property~\cite{bishop2006pattern}, which means that two inputs close in the feature space share similar contextual attributes.
For example, as shown in Figure \ref{fig:motivation}, there are some already labeled data surrounding the unlabeled data, and their classification results (correctly classified/misclassified) are already known. 
If an unlabeled instance is close to already falsely classified data, under the local continuity property, it is likely that this instance is also a model bug. 
This property enables us to extract the contextual attributes for an unlabeled instance from its neighboring labeled data. 
By combining the extracted contextual attributes together with the intrinsic attributes, we expect to achieve better bug-revealing capability estimation.



\section{Methodology}
\label{sec:metho}
In this section, we first provide an overview of the proposed method \textit{TestRank} (Section \ref{subsec:overview}). Then, we discuss the details of learning to extract the contextual attributes in Section \ref{subsec:gcn}. At last, we explain the combination of intrinsic and contextual attributes for final bug-probability estimation (Section \ref{subsec:collaborative}).  

\begin{figure*}[t]
     \centering
     \includegraphics[width=\linewidth]{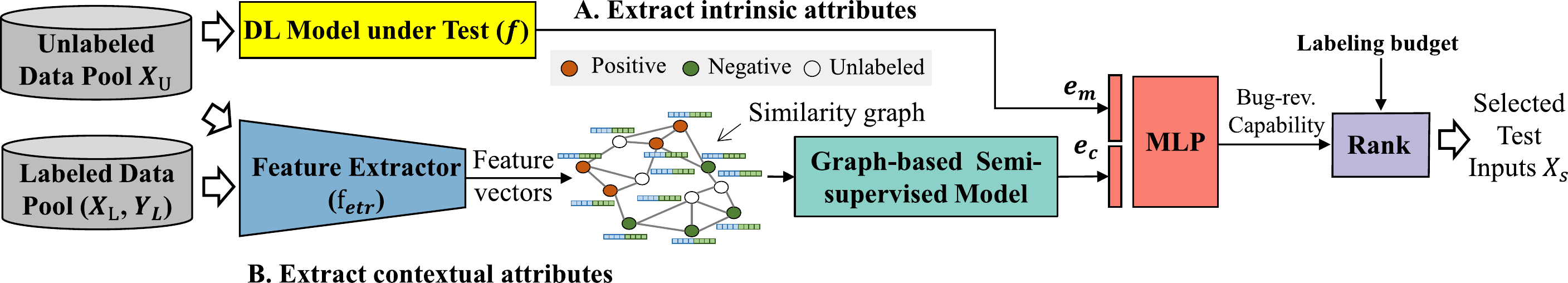}
     \caption{The overview of TestRank. For an input instance, path $A$ extracts intrinsic attributes from the target DL model. Path $B$ explores the relation between unlabeled test instances and the contextual information, i.e., labeled instances and their corresponding binary (positive/negative) labels from the target DL model, to extract the contextual attributes. Later, the intrinsic and contextual attributes are combined for collaborative bug-revealing probability estimation.}
     \label{fig:overflow}
\end{figure*}

\subsection{Overview}
\label{subsec:overview}

Figure \ref{fig:overflow} shows the overview of our method. TestRank consists of two attribute extraction paths for the final bug-revealing capability estimation: 
\begin{enumerate}
    \item \textbf{Path $A$: intrinsic attribute extraction}. Given a pool of unlabeled data pool $\mathbf{X}_U$, we use the DL model under test to extract the intrinsic attributes $\mathbf{e}_m$ for each input. More precisely, we collect the output logits (\textit{i.e.}, the vectors before $softmax$ layer) from the DL model $f$ as the intrinsic attributes $\mathbf{e}_m$. 
    \item \textbf{Path $B$: contextual attribute extraction}. Graph-based representation learning technique is applied on both unlabeled data pool $\mathbf{X}_U$ and labeled data pool  $\mathbf{X}_L$ (e.g. training set and historical test oracles). We use a well-trained feature extractor to map the original data space into a more confined feature space with good local continuity property. Then we create the similarity graph (\textit{i.e.,} $k$-Nearest Neighbor Graph) based on the obtained feature vectors. Last but not least, a graph neural network (GNN) model is designed to learn to extract the contextual attributes $\mathbf{e}_c$ of each instance from the similarity graph. The details are elaborated in Sec.~\ref{subsec:gcn}.
    \item \textbf{Bug-revealing capability estimation.} After attributes $\mathbf{e}_m$ and $\mathbf{e}_c$ are extracted, we combine and aggregate them together via a Multi-Layer Perceptron (MLP), which will be detailed in Section \ref{subsec:collaborative}. The outputs from MLP predict the final bug-revealing capability for each unlabeled test instance. At last, the unlabeled test instances are ranked according to their bug-revealing capability, and the top ones are selected under the given labeling budget.
\end{enumerate}

In the following subsections, we discuss the path $B$ and how to combine path $A$ and $B$ in detail.

\subsection{Contextual Attributes Extraction}
\label{subsec:gcn}

We represent the contextual information from the DL model as a set of labeled inputs $\mathbf{X}_L$ and the corresponding binary flags $\mathbf{Y}_L$, which indicate whether the DL model predicts each $\mathbf{x} \in \mathbf{X}_L$ correctly (flagged as $1$) or incorrectly (flagged as $0$). 
With this contextual information, our goal is to extract the contextual attributes for each instance in the unlabeled test instance set $\mathbf{X}_U$.

Extracting contextual attributes from labeled and unlabeled data is non-trivial. It is well-known that the real data, especially image data, is usually in high-dimensional space, wherein the underlying data distribution will live on complex and non-linear \textit{manifold} embedded within the high-dimensional space~\cite{bishop2006pattern}. Therefore, constructing the relationships between different instances is difficult. 
To address the challenge, we take a graph-based approach to capture the contextual attributes of test data.

To be specific, we first map the raw data into a confined feature space with the following local continuity property:

\textbf{Local Continuity Property of Feature Space.} \textit{Raw data is mapped into the feature space (the learned representation) where small changes will produce small changes in the output space. In other words, two inputs that are close in the feature space share similar contextual attributes.}

Then we propose to construct a similarity graph for both labeled and unlabeled data in the feature space. With the similarity graph, we can effectively aggregate the contextual information from adjacent labeled data to a specific instance with graph neural networks, which reveal the bug-revealing capability of this instance. 
The contextual attributes extracted by the graph neural network can then be combined with the intrinsic attributes to conduct the better bug-revealing capability estimation (See Section~\ref{subsec:collaborative}). The contextual attributes extraction process is formally depicted by Algorithm~\ref{algo:correlation}.

\begin{algorithm}[t]
\caption{GNN-based Contextual Attributes Extraction}
\label{algo:correlation}
\KwIn{Input samples $\mathbf{X}_U \cup \mathbf{X}_L$, Binary labels for labeled samples $\mathbf{Y}_L$, 
Number of neighbors $k$, Feature extractor $f_{etr}$, GNN layers $M$, number of training epochs.}

\KwOut{Contextual attributes $\textbf{E}_c$ for $\mathbf{X}_U$}
\tcc{{\color{red}Extract compact representation}}
$\mathbf{\overline X} = f_{etr}(\mathbf{X}_U \cup \mathbf{X}_L) $\;
\tcc{{\color{red}KNN Graph construction}}
$\mathbf{A, Edge} = knn\_graph(\mathbf{\overline X}, k)$\;
\tcc{{\color{red} Train GNN}}
$\mathbf{\tilde{A}} = \mathbf{Edge} + \mathbf{I}_{N}$\;
$\mathbf{\tilde{D}} = \sum_j{\mathbf{\tilde{A}}_{i,j}}$\;
$\mathbf{H}^0 = \mathbf{\overline X}$\;
\For{Number of training epochs}
{
\For{$l=0, 1, \ldots, M$}
{
$    \mathbf{H}^{l+1} = \sigma(\mathbf{\tilde{D}}^{-\frac{1}{2}}\mathbf{\tilde{A}}\mathbf{\tilde{D}}^{-\frac{1}{2}}\mathbf{H}^{l}\mathbf{\Theta}^{l}),
$
}
$Output = \text{FCLayer} (\mathbf{H}^{M+1})$\;
$loss = \text{CrossEntropyLoss}(Output, \mathbf{Y}_L)$\;
Back propagation\;
Update $\mathbf{\Theta}$\;
}
$\mathbf{E}_c = \mathbf{H}^{M+1}$[unlabeled\_index]\;
\Return $\mathbf{E}_c$;
\end{algorithm}

\vspace{3pt}
\noindent \textbf{Feature Vector Representation (Line 1).}
As we focus on the image classification task, the original high-dimensional input space has poor local continuity property, as the simple distance metrics (\textit{e.g.}, Euclidean, Cosine) fail to measure the similarity between inputs. Therefore, mapping the original image space into another feature space with good local continuity property is necessary for similarity measurement.

Unsupervised learning is a way to learn compact and confined representations from the original high dimensional data without labels \cite{Bastien2020Bootstrap, orrd2018representation}.
Among the unsupervised learning techniques, the BYOL \cite{Bastien2020Bootstrap} model explicitly introduces local continuity constraint into learned feature space and shows the state-of-the-art results on various downstream tasks. 
Therefore, we employ a BYOL model, denoted by $f_{etr}$, to extract the features from the input images.
The data used to train the BYOL model includes input instances from both labeled and unlabeled data pool: $(X_U \cup X_L)$. Please note that the feature extractor can be replaced by any other feature extractor models with the local continuity property and good performance, like SimCLR~\cite{chen2020simple} and MoCo~\cite{he2020momentum}.

\vspace{3pt}
\noindent \textbf{Similarity Graph Construction and Approximation (Line 2).} 
After the extraction of feature representation, we use the Cosine distance metric to measure the similarity between any two test instance $i$ and $j$:
\begin{equation}
    Dist(i,j) = \text{Cosine}(f_{etr}(\mathbf{x}_i), f_{etr}(\mathbf{x}_j)), \quad \mathbf{x}_{i}, \mathbf{x}_{j} \in \mathbf{X}_L \cup  \mathbf{X}_U.
\end{equation}

Based on the distance matrix $Dist$, we construct a $k$-NN Graph $\mathcal{G}$, wherein each sample is connected to its top-$k$ most similar samples. The connection in the similarity graph is represented by an adjacency matrix $\mathbf{A} \in \mathcal{R}^{N\times N}$, where $N$ is the number of sample in $\mathbf{X}_L \cup \mathbf{X}_U$. The entry $\mathbf{A}_{ij}$ equals 1 if node $j$ is within the $k$ nearest neighbors of node $i$, and 0 otherwise. The edge weight matrix of the similarity graph is denoted as $\mathbf{Edge}$, wherein each edge weight in $\mathbf{Edge}$, if exists, is inversely proportional to the corresponding distance $Dist$:
\begin{equation}
    \mathbf{Edge}_{ij} = 
            \begin{cases}
            1/Dist(i,j)& {\mathbf{A}_{ij} = 1.}\\
            0& {\mathbf{A}_{ij} = 0.} 
            \end{cases}
            \quad i, j \in \{0,\ldots,N-1\}.
\end{equation}
This means that the connection between two nodes, if exists, is weaker if their proximity is large.

Constructing such a $k$-NN graph is, however, computationally expensive. To calculate the distance between each pair of test instances, the required computational complexity is $O(N^2)$, which is prohibitive to scale up the current massive unlabeled test instances in real applications. Therefore, we propose an approximation method for $k$-NN graph construction. Our intuition is that, since the aim of graph construction is to exploit the bug patterns of the nearby labeled instances for the unlabeled instances, the connections between unlabeled data are less important. Therefore, we propose to only consider the connections among labeled data $\mathbf{X}_L$, and the connections between labeled $\mathbf{X}_L$ to unlabeled data $\mathbf{X}_U$. This approximation reduces the cost from $O(N^2)$ to $O(P^2 + PQ)$, where $P$ and $Q$ stand for the number of data in $\mathbf{X}_L$ and $\mathbf{X}_U$, respectively. Usually, in the real-world scenario, $P$ is much smaller than $Q$, thereby we could obtain a near-linear graph construction algorithm with complexity $O(PQ)$.

\vspace{3pt}
\noindent
\textbf{GNN-based representation Learning (Line 3-15).}
Graph neural network (GNN) is a simple and effective way to aggregate neighboring information and learn local patterns \cite{KipfW17}.
After the construction of graph $\mathcal{G}$, as the classification results of $\mathbf{X}_L$, \textit{i.e.}, $\mathbf{Y}_L$, are already known, we apply the semi-supervised GNN algorithm on the similarity graph to extract the contextual attributes for the unlabeled test instances.
To be specific, we use an auxiliary binary classifier, which is supervised on the labeled data, to learn the GNN model parameters.

To apply the GNN algorithm, we first initialize the input feature vector $\mathbf{H}^0$ of each node in graph $\mathcal{G}$:
\begin{equation}
    \mathbf{H}^0_{i} = f_{etr}(\mathbf{x}_i),\quad \mathbf{x}_i \in \mathbf{X}_U \cup \mathbf{X}_L.
\end{equation}
Recall that in each GNN layer, the feature vectors are propagated between neighbors and aggregated together. Thus, we can obtain the features in the next GNN layer by:
\begin{equation}
    \mathbf{H}^{l+1} = \sigma(\mathbf{\tilde{D}}^{-\frac{1}{2}}\mathbf{\tilde{A}\tilde{D}}^{-\frac{1}{2}}\mathbf{H}^{l}\mathbf{\Theta}^{l}),
\end{equation}
where $\mathbf{\tilde{A}} = \mathbf{Edge} + \mathbf{I}_{N}$, $\mathbf{I}_{N}$ is the identity matrix, $\mathbf{\tilde{D}} = \sum_j{\mathbf{\tilde{A}}_{i,j}}$, $\mathbf{\Theta}^{l}$ is the trainable weight matrix for the $l^{th}$ layer, $\sigma$ is an activation function and output feature vector matrix. 
The propagation and aggregation are repeated for $M$ GNN layers. Then, the output feature vector $\mathbf{H}^{M}$ is processed by a fully connected layer to generate the final binary classification results.

Then, for any labeled node in $\mathbf{X}_L$, we could obtain a binary cross entropy between the calculated binary label and the expected binary label:
\begin{equation}
  \mathcal{L}_{ce} = \sum y log(\text{GNN}(\mathbf{x}_e)), \quad \forall \mathbf{x}_e \in  f_{etr}(\mathbf{X}_L), \mathbf{y} \in \mathbf{Y}_L
\end{equation}
where $\text{GNN}(\mathbf{x})$ denotes the GNN model. The GNN model is trained via minimizing the loss for some training epochs (we set it as 600 in our experiment).
After that, we apply the trained GNN model on $\mathbf{X}_U$ and to obtain the $\mathbf{E}_c$ (line 15).

\subsection{Bug-revealing Capability Estimation}
\label{subsec:collaborative}
Finally, we combine both the intrinsic attributes and contextual attributes for collaborative bug-capability inference. To achieve a proper collaboration, we formulate the combination function as a neural network. 
Specifically, we concatenate the intrinsic $\mathbf{e_m}$ and contextual attributes $\mathbf{e_c}$ for a test instance as the input to the neural network.
Since these two vectors are condensed vectors, we could use an MLP to handle them,  albeit other types of networks are applicable. 

\begin{equation}
\label{eq:output}
    \begin{split}
    Output = \text{MLP}(\text{concat}(\mathbf{e_c}, \mathbf{e_m})), \\
    Bug \ Rev. \ Capability = S(output)
    \end{split}
\end{equation}

The MLP is trained with the labels instances $(\mathbf{X}_L, \mathbf{Y}_L)$ in a supervised manner. The objective is to minimize the corresponding binary Cross-Entropy loss function. The final bug-revealing probability of unlabeled data is produced by the learned MLP by applying a $sigmoid$ function $S(t)={\frac {1}{1+e^{-t}}}$ on the $output$ for an unlabeled instance (see Equation \ref{eq:output}).  
After training, the MLP shall re-weight the importance of intrinsic and contextual attributes, and make a final decision by assigning a high probability to a test instance if it is likely to trigger a bug.

Finally, we rank the unlabeled test instances in a descending order based on their bug revealing capability. Under the given budget, we select the top ones to label and test.  

%% file: result.tex
\section{Experiment}
\label{sec:experiment}

\begin{table*}[t]
\caption{The Dataset and DL models.}
\label{table:dataset}
\resizebox{\textwidth}{!}{  
\begin{tabular}{ccccccc}
\toprule
Dataset  & \# Class & Dataset Size & Official Train/Test/Extra Split & Our Data Split(TC/DC/HO) & Model Architecture & Model Acc. On HO set(\%) (A/B/C) \\ \midrule
CIFAR-10 & 10       & 60K          & 50K/10K/x                       & 20K/39K/1K           & ResNet-18          & 70.1/66.4/68.3                  \\
SVHN     & 10       & 630K         & 73K/26K/531K                    & 50K/49K/531K         & Wide-ResNet        & 94.2/92.5/81.6                  \\
STL10    & 10       & 13K          & 5K/8K/x                         & 5K/7.5K/0.5K         & ResNet-34          & 54.8/54.0/53.6                  \\ \bottomrule
\end{tabular}
}
\end{table*}

In this section, we present our experimental setup in Section \ref{subsec:setup}. Then, we discuss the debugging effectiveness of the proposed method in Section \ref{subsec:debugeffectiveness}. At last, we discuss the impact of hyper-parameters on the performance of our method in Section \ref{subsec:hyper}.

\subsection{Setup}
\label{subsec:setup}

\textbf{Datasets.}
We evaluate the performance of our method on three popular image classification datasets: CIFAR-10~\cite{krizhevsky2009learning}, SVHN~\cite{Netzer2011ReadingDI}, and STL10~\cite{Coates2011AnAO}. 
The details of them are shown in Table \ref{table:dataset}. 
CIFAR-10 is officially composed of 50,000 training images and 10,000 test images, and it has ten classes of natural images. The Street View
House Numbers (SVHN) dataset contains house numbers from Google Street View images. It contains 73,257 training images and 26,032 testing images. Besides, the SVHN dataset also has an extra set of 531,131 images.
The STL10 dataset contains ten classes of natural images. In each class, there are 500 training images and 800 test images.

There are mainly two parties involving in the model construction process: the training center and the debugging center. Hence, we manually split the dataset into the training center dataset (see the \textit{TC} column in Table \ref{table:dataset}) and the debugging center dataset (see the \textit{DC} column in Table \ref{table:dataset}). To mimic the practical scenario where the unlabeled data is abundant, we let the debugging center owns more data. In the debugging center, we let a set of test data as labeled ones to represent the historical test oracles, and they are 8K/10K/1.5K for CIFAR-10, SVHN, and STL10, respectively, which are used to train the GNN and MLP model. The rest of the data in the debugging center are left unlabeled. Also, we spare a hold-out dataset (see the HO column), which is used for evaluating the accuracy.

\vspace{3pt}
\noindent
\textbf{Target DL model (model under test).}
We use the popular ResNet and WideResNet architectures as the backbone models \cite{He2016DeepRL, Zagoruyko2016WRN}, due to their state-of-the-art performance on various datasets. The ResNet version used for each dataset is shown in Table \ref{table:dataset}.
To simulate DL models of different qualities, for each dataset, we train three DL models with three sub-sets drawn from the training set owned by the training center. Each sub-set is randomly drawn from the training set. For model B and C, the training set are drawn with in-equivalent class weights. Then, we further split the selected sub-set to train and validation set for DL model training. After training, we report the accuracy of models on the debugging center's hold-out dataset $T_{HO}$ in Table \ref{table:dataset}.

Our code is implemented with open-source PyTorch and PyG \cite{Fey2019FastGR} machine learning libraries. All of our experiments are performed on a single TITAN V GPU. Besides, we will release our code upon acceptance.

\vspace{3pt}
\noindent
\textbf{Baselines.} Given a DL model and a certain budget, the goal of our method is to select test cases from an unlabeled data pool to discover the bugs of the given DL model. We compare our work with the following representative test selection techniques:
\begin{itemize}
    \item \textbf{DeepGini \cite{Feng2020DeepGini}}: DeepGini is the state-of-art test case selection technique. Given a model under test, DeepGini ranks test cases to be labeled by a score defined based on the output confidence (See Equation \ref{eq:confidence} for the definition). 
    \item \textbf{MCP \cite{Shen2020MCP}}: In addition to the output probabilities of the model under test, MCP also considers the balance among different class boundaries of the selected test inputs. Specifically, they group test cases into different clusters, where each cluster stands for a distinct classification boundary, and try to equally choose prioritized test cases from each cluster.
    \item \textbf{DSA \cite{ByunSVRC19}}: 
    Byun \textit{et. al.} propose to use the distance-based surprise score (DSA) as a test prioritization metric, which is originally proposed in \cite{Kim2019GuidingDL}. The surprise score measures how far the test case w.r.t. the training set. Samples with higher surprise scores are preferred for testing. Please refer to \cite{ByunSVRC19} for a detailed calculation.
    \item \textbf{Uncertainty \cite{ByunSVRC19}}: Besides the surprise score, Byun \textit{et. al.} also propose to use the dropout uncertainty to rank test inputs. To be specific, the dropout uncertainty is calculated by running the model multiple times (e.g. $t$ times) with a certain dropout rate. Then, the output probability is averaged among the $t$ times. The final uncertainty is calculated as the entropy on the averaged output probabilities.
    \item \textbf{Random:} Test inputs are randomly drawn from all unlabeled samples. 
\end{itemize}

\noindent
\textbf{Evaluation metric.} 
We propose a new performance measurement metric for test prioritization techniques: \textit{TPF}. \textit{TPF} is defined as the number of detected bugs divided by the number of budget or the number of total bugs identified by the whole unlabeled test set, whichever is minimal:

\begin{equation}
\label{eq:tpf}
     TPF = \frac{\# Detected \ Bugs}{MIN(\# Budget, \# Total \ Bugs)}.
\end{equation}
When \# budget $\leq$ \# total bugs, the maximum number of bugs can be identified by a test prioritization technique equals to the budget. When \# budget $\geq$ \# total bugs, the maximum number of bugs can be detected equals to the total number of bugs. Therefore, \textit{TPF} measures how far a test prioritization technique is to the ideal case.

In practice, under the massive unlabeled data, the selection performance under a small budget is considered more important than that under a large budget. 
Therefore, to provide an insight on the quality of the test prioritization technique under a small budget, we also provide an ATPF metric: ATPF measures the average TPF values for a budget less than the total bugs:
\begin{equation}
    ATPF = \frac{1}{N} \sum_i^{N-1} TPF_i,
\end{equation}
where $TPF_i$ stands for the $i$-th TPF value under budget $b_i, i \in \{0,\ldots,N-1\}$, and $b_i \leq$ number of total bugs. 



The proposed metrics enhance the ones used by Feng \textit{et. al.} \cite{Feng2020DeepGini} and Byun \textit{et. al.} \cite{ByunSVRC19}. They use the percentage of detected bugs against the percentage of budget (and an APFD \cite{Do2006OnTU} value which is devised based on it) for evaluation. This metric would produce a small value under a small budget, regardless of how good the prioritization technique is.
For example, we assume that there are 10,000 unlabeled data and 2,000 of them can detect model bugs. If the budget is 100, the best percentage of detected bugs is 5\%, and the worst is 0\%. Under this metric, the gap between the best and the worst is only 5\%. By contrast, TPF enlarges this gap to 100\% to better differentiate the ability of different test prioritization techniques.




\begin{table*}[t]
\caption{Comparison of \textit{TextRank} with baseline methods in terms of the average debugging effectiveness (ATPF values (\%)). Except for random selection, all baselines only use the intrinsic attributes of unlabeled instances for bug-revealing capability estimation. In contrast, \textit{TestRank} leverages both the intrinsic and contextual attributes.  }
\resizebox{0.7\textwidth}{!}{  
\begin{tabular}{ccccccccc}
\toprule
Dataset                   & Model ID &                          Random  & MCP   & DSA   & Uncertainty & DeepGini  & \begin{tabular}[c]{@{}c@{}}TestRank\\ Contextual-Only\end{tabular} & TestRank           \\ \toprule
\multirow{3}{*}{CIFAR-10} & A & 30.15   & 58.25 & 60.93 & 58.09       & 67.47  & 51.39  & \textbf{76.56} \\
                          & B & 34.18   & 46.46 & 62.34 & 61.85       & 67.80  & 58.85  & \textbf{87.87} \\
                          & C & 34.27   & 65.25 & 64.47 & 63.10       & 71.15  & 75.33  & \textbf{85.53} \\ \midrule
\multirow{3}{*}{SVHN}     & A & 10.16   & 39.98 & 55.47 & 58.29       & 63.47  & 44.16  & \textbf{66.06} \\
                          & B & 11.85   & 38.07 & 57.96 & 58.06       & 63.85  & 51.26  & \textbf{76.36} \\
                          & C & 23.41   & 65.33 & 69.34 & 71.80       & 81.68  & 93.99  & \textbf{95.32} \\ \midrule
\multirow{3}{*}{STL10}    & A & 39.25   & 66.62 & 64.56 & 64.30       & 69.70  & 60.09  & \textbf{79.00} \\
                          & B & 42.60   & 69.97 & 67.12 & 65.30       & 72.89  & 71.90  & \textbf{80.96} \\
                          & C & 46.05   & 71.88 & 66.60 & 70.34       & 73.34  & 79.55  & \textbf{88.67} \\ \bottomrule
\end{tabular}
}
\label{table:debug_effectiveness}
\end{table*}

\subsection{Comparison of TestRank with Baselines}
\label{subsec:debugeffectiveness}

In this section, we evaluate \textit{TestRank} against five baseline methods (\textit{e.g.}, Random, DeepGini, MCP, DSA, and Dropout-uncertainty) with the \textit{TPF} and \textit{ATPF} metrics. For the dropout uncertainty method, we run 1000 times inferences with a dropout rate of 0.5. For the DSA method, we use the final convolution layer to collect the activation traces to calculate the surprise score. For our method, we set the number of neighbors for constructing the kNN graph as 100. Also, we use a two-layer GNN with a hidden dimension 32.

Table \ref{table:debug_effectiveness} compares \textit{TestRank} with baselines w.r.t the overall debug effectiveness with the \textit{ATPF} metric. 
From this table, we have several observations. 
First, comparing with the baselines, \textit{TestRank} can achieve the highest ATPF values on all evaluated datasets and models. 
For instance, on CIFAR-10, \textit{TestRank} can achieve 9.09\%, 20.07\%, 14.38\% higher ATPF values than the best baseline DeepGini for model A, B, C, respectively. 
Therefore, our method can distinguish the bug-revealing capability of the unlabeled test inputs much more accurately than baselines. 
Second, the \textit{TextRank-Contextual-Only} column shows the result using only the contextual attributes. 
We observe that the contextual attributes alone can achieve higher debug effectiveness than random prioritization. 
For example, for model A on CIFAR-10, the effectiveness of random prioritization is 30.15\% while that of the context-only method is 51.39\%.
We manually check the distribution of bugs of model C and find that many bugs are centralized on two classes, where the training data is insufficient.
This kind of bug is easily detected by the contextual attributes-based method.
Hence, the contextual information is a useful one to improve test prioritization effectiveness. 
But still, the context attributes alone are not sufficient. 
The combination of intrinsic and contextual attributes is essential in achieving high accuracy bug-revealing capability estimation.

\begin{figure*}
	\includegraphics[width=0.95\linewidth]{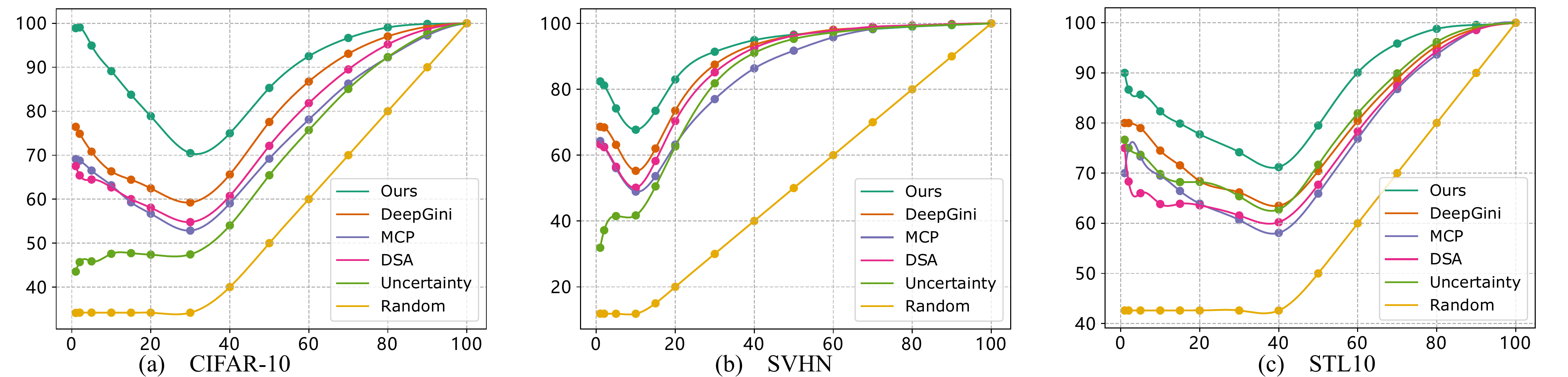}
	\caption{The debugging effectiveness (TPF values) against all budgets. X-axis: the budget measured in percentage. Y-axis: the TPF value. Please note that this figure is generated with model B on each dataset.}
	\label{figs:perf_all_budget}
\end{figure*}

To show more detailed results, we present \textit{TPF} value for each target DL model against every labeling budget in Figure \ref{figs:perf_all_budget}. 
The $x$-axis represents the labeling budget measured in percentage, and the $y$-axis is the \textit{TPF} values.
As we can observe, the TPF values for most curves decrease in the beginning and then increase. 
The turning point is when the budget equals to the number of total bugs triggered by the whole unlabeled set, i.e., \# budget $=$ \# total bugs. 
When \# budget $<$ \# total bugs, the reason why the TPF values are decreased is that we rank the test instances according to their bug-revealing probability. Assuming the bug-revealing probability estimation is accurate, with the increase of budget, the selected test cases will have a lower average bug-revealing ability, thus the decreased TPF value.
When \# budget $>$ \# total bugs, according to the definition of TPF (see Equation \ref{eq:tpf}), the denominator is fixed. Since the increase of budget will increase the number of detected bugs, the TPF value will increase.

Figure \ref{figs:perf_all_budget} shows that our method constantly outperforms other methods. The improvement of our method is especially obvious when the budget is small. For example, in Figure \ref{figs:perf_all_budget} (a), \textit{TestRank} improves the debug efficiency by around $20\%$ compared to the best baseline when the budget is around $1\%$. When the budget is rather high (e.g. budget $\geq 80\%$), the difference between different methods is less obvious, because most bugs can be selected under the large budget.

To sum up, TextRank shows considerable improvements in the debug effectiveness against baselines, and the improvements are more significant when the budget is small.


\subsection{Influence of TestRank Configurations}
In this section, we investigate the impact of different \textit{TestRank} configurations.
\label{subsec:hyper}

\begin{table}[]
\caption{The performance (ATPF values) of TestRank under different configurations.}
\resizebox{0.48\textwidth}{!}{
\begin{tabular}{cccccc}
\toprule
Dataset                   & Model & \begin{tabular}[c]{@{}c@{}}TestRank \\ (\%)\end{tabular} & \begin{tabular}[c]{@{}c@{}}TestRank \\  w/o approx. (\%)\end{tabular} & \begin{tabular}[c]{@{}c@{}}TestRank\\ TargetModel(\%)\end{tabular} \\ 
\toprule
\multirow{3}{*}{CIFAR-10} & A        & 76.56         & 77.77 (+1.21)                    & 68.84 (-7.71)                            \\
                          & B        & 87.87         & 87.70 (-0.17)                    & 81.46 (-6.40)                            \\
                          & C        & 85.53         & 88.10 (+2.57)                    & 77.73 (-7.79)                             \\ \midrule
\multirow{3}{*}{SVHN}     & A        & 66.06         & 63.87 (-2.19)                    & -                                    \\
                          & B        & 76.36         & 82.04 (+5.68)                    & -                                     \\
                          & C        & 95.32         & 96.62 (+1.30)                    & -                                     \\ \midrule
\multirow{3}{*}{STL10}    & A        & 79.00         & 80.50 (+1.50)                    & 67.59 (-11.40)                        \\
                          & B        & 80.96         & 78.98 (-1.98)                    & 74.43 (-6.52)                          \\
                          & C        & 88.67         & 89.32 (+0.65)                    & 72.43 (-16.23)                         \\ \midrule
\multicolumn{3}{c}{Average Influence (\%)}           & \textbf{+0.95}            & \textbf{-6.23}                       \\ \bottomrule
\end{tabular}
}
\label{table:configurations}
\end{table}

\vspace{3pt}
\noindent
\textbf{Feature Extractor.}
\textit{TextRank} uses an unsupervised BYOL model trained on both labeled and unlabeled data to extract their features. One may wonder if the unsupervised model can be replaced by a supervised model (\textit{e.g.}, the target DL model).
To investigate this, we replace the feature extractor in \textit{TestRank} with the front layers (we remove the last few linear layers) of the target DL model.
The result is shown in the \textit{TextRank-TargetModel} column in Table \ref{table:configurations}. 
We do not report the result on the SVHN dataset because the size of the intermediate features obtained by the WideResNet is too large, and constructing $k$-NN graph on large vectors is time-consuming.
Comparing with the original \textit{TextRank}, the average ATPF value on the reported datasets and models reduces by 6.23\%, which is significant.  
As the quality of the target DL model given to the debugging center is unknown, its feature extraction performance is not reliable. 
Also, the unsupervised model uses the unlabeled data during training, which enables it to extract better features for these data. 
Therefore, the debugging center must train a separate unsupervised feature extractor for better performance.


\vspace{3pt}
\noindent
\textbf{$k$-NN graph approximation.}
To reduce the computation complexity, TextRank uses approximation techniques when constructing the $k$-NN graph (see Section \ref{subsec:gcn}). The \textit{TextRank-w/o-approx.} column in Table \ref{table:configurations} shows the debug effectiveness when we use the original $k$-NN graph construction without approximation.
The result indicates that the average influence of the approximation is 0.95\%, which is small. 
If the computation allows or the unlabeled data set is small, one can resort to the original $k$-NN graph construction for better performance. 
However, if there is a computation resource limit and the unlabeled test instances are massive, we recommend using the approximation to greatly save computation with negligible performance loss.

\begin{figure}
    \centering
    \includegraphics[width=0.85\linewidth]{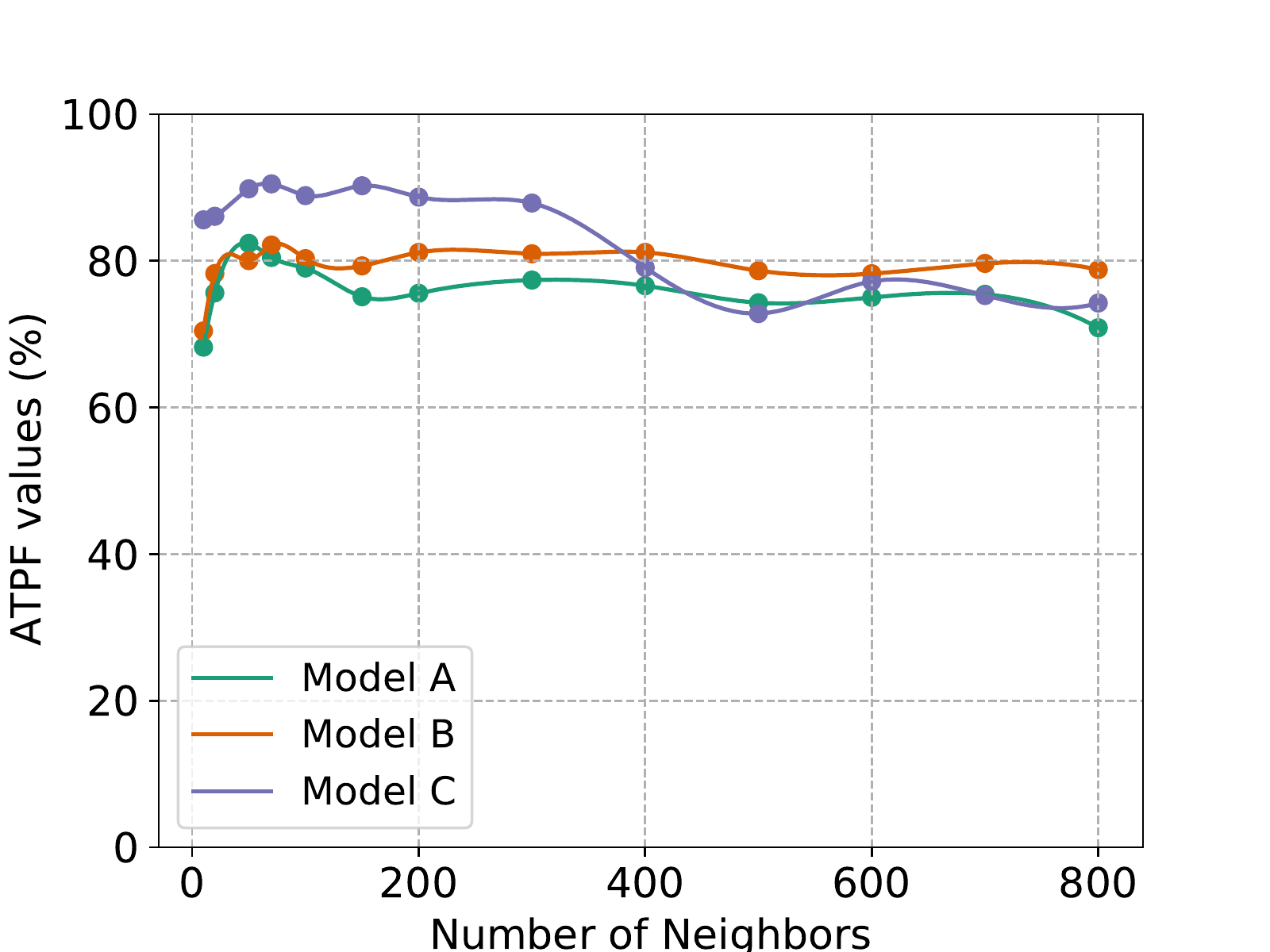}
    \caption{The impact of the number of neighbors $k$ on the debug effectiveness (The experiments are performed on STL10 dataset). }
    \label{fig:neighbors}
\end{figure}

\vspace{3pt}
\noindent
\textbf{Number of nearest neighbors $k$.}
When constructing the $k$-NN graph, the number of nearest neighbors $k$ decides the range of the context one node/instance can reach.
In previous experiments, the $k$ is set to 100. We enlarge this range to (20 - 800) to study the influence of $k$. The result is shown in Figure \ref{fig:neighbors}. 
One can observe that the debug effectiveness will decrease either when $k$ is too small or too large. When $k$ is too small, the context information available to one instance is limited, which makes it difficult for the GNN to extract useful contextual attributes. When $k$ is too large, the GNN may grasp some irrelevant/noisy information. 
We also observe that, \textit{TextRank} can achieve good performance in a wide range of $k$ values. For example, for model A, TextRank is better than the best baseline 69.70\% (see Table \ref{table:debug_effectiveness}) when $k$ is larger than 20.
But still, setting a proper $k$ is necessary in obtaining optimized debugging performance. We leave this as our future work.

Overall, from these analyses, we have the following conclusions. 
First, using a separate unsupervised feature extraction model is necessary. Second, approximating the $k$-NN graph construction can save a lot of computation without much performance loss when the unlabeled data is massive. Third, selecting the number of nearest neighbors $k$ is relatively flexible, as TextRank produces stable performance on a wide range of $k$. 





%% file: conclusion.tex


\section{Conclusion and Future Work}
\label{sec:conclusion}
We propose \textit{TestRank}, a novel test prioritization framework for DL models. 
To estimate a test instance's bug-revealing capability, \textit{TextRank} not only leverages the intrinsic attributes of an input instance obtained from the target DL model, but also extracts the contextual attributes from the DL model's historical inputs and responses (\textit{e.g.}, the training instances and their correctness assigned by the DL model). Our empirical results show that \textit{TestRank} can capture the bug-revealing capabilities of unlabeled test instances more accurately than existing solutions.

In this paper, we consider each test case equally and aim to identify as many bug-revealing test cases as possible. In practice, the impact of these test cases could be different.  We leave the study for such impact for future work. Besides, the current \textit{TestRank} solution only supports CNN-based classification tasks, and we plan to extend it to other tasks with different DL models (\textit{e.g.}, RNN and GNN models) in the future.